\documentclass[conference]{IEEEtran}
\IEEEoverridecommandlockouts
\usepackage{cite}
\usepackage{comment}
\usepackage{amsmath,amssymb,amsfonts}
\usepackage{algorithm}
\usepackage{algorithmic}
\usepackage{graphicx}
\usepackage{textcomp}
\usepackage{xcolor}
\usepackage{float}
\usepackage{multirow}
\usepackage{adjustbox}
\usepackage{cleveref}
\usepackage{subcaption}
\usepackage{flushend}
\usepackage[caption=false]{subfig}
\def\BibTeX{{\rm B\kern-.05em{\sc i\kern-.025em b}\kern-.08em
    T\kern-.1667em\lower.7ex\hbox{E}\kern-.125emX}}
\begin{document}

\title{Accuracy-Privacy Trade-off in the Mitigation of Membership Inference Attack in Federated Learning}

\author{
    \IEEEauthorblockN{
    Sayyed Farid Ahamed\IEEEauthorrefmark{1},
    Soumya Banerjee\IEEEauthorrefmark{1},
    Sandip Roy\IEEEauthorrefmark{1},
    Devin Quinn\IEEEauthorrefmark{2},  
    Marc Vucovich\IEEEauthorrefmark{2},
    }
    \IEEEauthorblockN{ 
    Kevin Choi\IEEEauthorrefmark{2}, 
    Abdul Rahman\IEEEauthorrefmark{2},
    Alison Hu\IEEEauthorrefmark{2},
    Edward Bowen\IEEEauthorrefmark{2},
    Sachin Shetty\IEEEauthorrefmark{1}
    }
    \IEEEauthorblockA{
    \IEEEauthorrefmark{1}Center for Secure \& Intelligent Critical Systems, Old Dominion University, Virginia, USA
    \\\{saham001, s1banerj, sroy, sshetty\}@odu.edu}
    \IEEEauthorblockA{
    \IEEEauthorrefmark{2}Deloitte \& Touche LLP
    \\\{devquinn, mvucovich, kevchoi, abdulrahman, aehu, edbowen\}@deloitte.com}
}

\maketitle

\begin{abstract}
Over the last few years, federated learning (FL) has emerged as a prominent method in machine learning, emphasizing privacy preservation by allowing multiple clients to collaboratively build a model while keeping their training data private. Despite this focus on privacy, FL models are susceptible to various attacks, including membership inference attacks (MIAs), posing a serious threat to data confidentiality. In a recent study, Rezaei \textit{et al.} revealed the existence of an accuracy-privacy trade-off in deep ensembles and proposed a few fusion strategies to overcome it \cite{rezaei2023accuracy}. In this paper, we aim to explore the relationship between deep ensembles and FL. Specifically, we investigate whether confidence-based metrics derived from deep ensembles apply to FL and whether there is a trade-off between accuracy and privacy in FL with respect to MIA. Empirical investigations illustrate a lack of a non-monotonic correlation between the number of clients and the accuracy-privacy trade-off. By experimenting with different numbers of federated clients, datasets, and confidence-metric-based fusion strategies, we identify and analytically justify the clear existence of the accuracy-privacy trade-off.
\end{abstract}


\section{Introduction}

Over the past few years, federated learning (FL) has gained significant traction as a widely adopted method in the field of machine learning (ML) with a primary focus on preserving privacy. FL enables multiple clients to work together in building a shared and cohesive model, all the while safeguarding the privacy of their individual training data \cite{mcmahan2017communication}.
Although FL is intended to protect individual's personal data,
recent studies have revealed that FL models can be susceptible to various attacks that expose sensitive information from their training datasets, including source inference, model inversion, and reconstruction attacks \cite{banerjee2023miabad}.
\begin{figure}
    \centering
    \includegraphics[width=\linewidth]{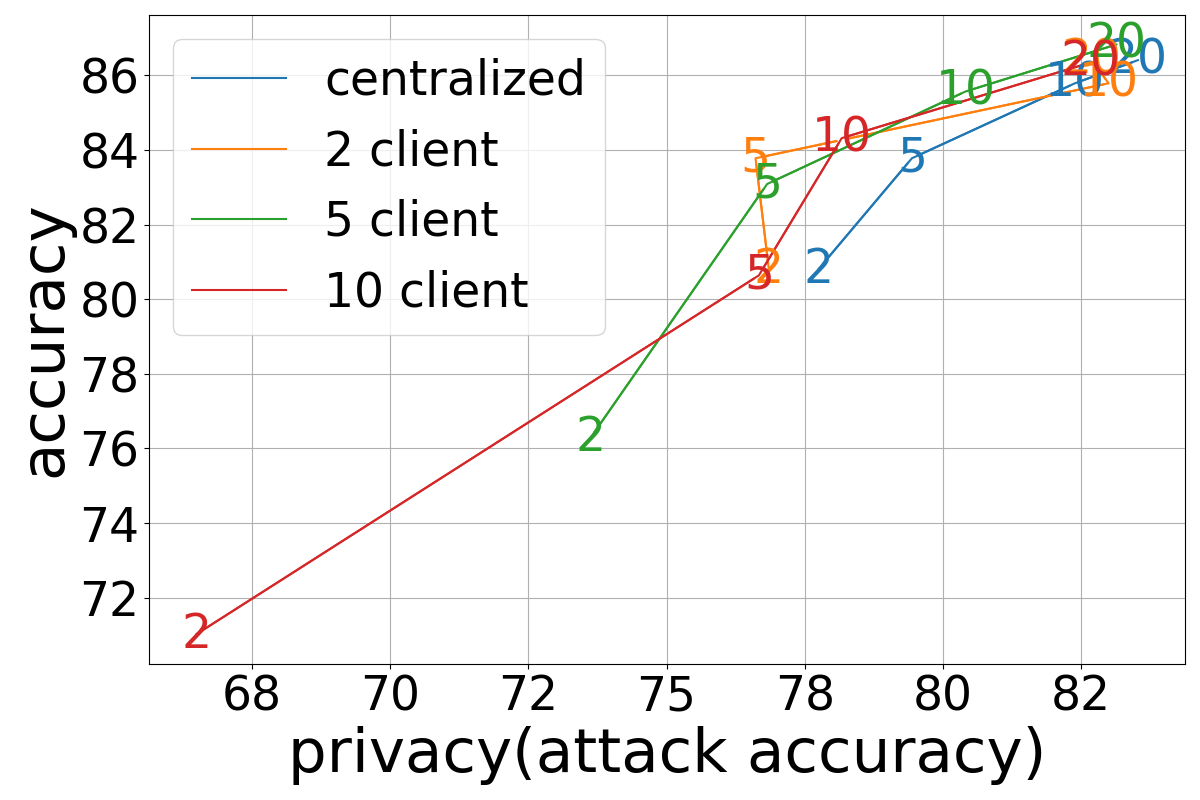}
    \caption{\textbf{Accuracy-privacy correlation in FL:} Training EfficientNet on CIFAR10. Each curve illustrates the evolution of accuracy and privacy over the training period. Notably, the test accuracy consistently rises while privacy decreases, demonstrating independence from the number of FL clients.}
    \label{fig:one}
\end{figure}

Membership inference attacks (MIAs) are indeed considered a serious threat to FL, as they can compromise the privacy and confidentiality of participant data by revealing whether specific samples were part of the training dataset, potentially undermining the security and trustworthiness of the framework. The accuracy of the MIA is widely accepted as a de-facto measure of an ML model's privacy \cite{rezaei2023accuracy}.
In 2020, Nasr \textit{et al.} showcased that in the realm of FL, adversarial participants can adeptly carry out active MIA against other participants, even in situations where the global model attains a high degree of predictive accuracy \cite{nasr2019comprehensive}.

FL and deep ensemble learning are two complementary approaches, both characterized by a collaborative nature, where multiple models or learners cooperate to enhance predictive performance in ML classifiers. FL addresses the challenges of privacy and decentralization by allowing multiple devices or servers to collaboratively train a global model without sharing raw data, while deep ensemble learning focuses on improving model performance and reliability by combining predictions from diverse, independently trained models. FL prioritizes privacy in a decentralized setting, with clients collaborating on a global model while keeping data locally. In contrast, ensemble learning focuses on improving accuracy by combining independently trained models without privacy concerns. The key difference lies in their primary objectives: privacy preservation for FL and enhanced predictive performance for ensemble learning. 

In 2022, S. Rezaei has shown that the effectiveness of MIA increases when ensembling improves accuracy \cite{rezaei2023accuracy}. Their empirical study indicates that this accuracy-privacy trade-off exists even for more advanced and state-of-the-art ensembling techniques. Interestingly, they break this trade-off by changing the fusing mechanism of deep ensembles which improves accuracy and privacy simultaneously. Instead of calculating the average of confidence values, their approach provides the confidence score of the most confident model within the ensemble that predicts the same label.

In this paper, \textbf{we aim to investigate two fundamental research inquiries} by exploring the inherent connection between deep ensembles and FL. Firstly, \textit{``With respect to MIA, does there exist an accuracy-privacy trade-off in a FL environment?".} Secondly, \textit{``Do the various confidence-based metrics adopted to break the trade-off for deep-ensemble learning also hold for FL?"}


In \Cref{fig:one}, we present a visual representation of the accuracy-privacy dynamics in FL. Specifically, the figure showcases the training of EfficientNet \cite{tan2019EfficientNet} on CIFAR10 \cite{krizhevsky2009learning}, where each curve depicts the evolution of accuracy and privacy with rounds. Notably, the curves illustrate a consistent rise in accuracy coupled with a decrease in privacy, revealing the independence of these dynamics from the number of FL clients. 

The main contributions of this paper, building upon the findings of \Cref{fig:one}, are as follows:

\begin{itemize}
    \item We demonstrate that within an FL framework, there exists a strong correlation between model accuracy and privacy, particularly in the context of MIA. This correlation underscores a clear and inherent trade-off between model accuracy and privacy in the FL setting.

    \item We make empirical studies with different numbers of clients for various datasets in a federated setting and establish that the number of clients is not monotonically correlated with accuracy and privacy.

    \item We implement various confidence-metric-based fusion strategies, previously used in deep ensembles, to enhance accuracy and privacy. However, we identify an accuracy-privacy trade-off in the FL environment and provide analytical justification for its presence.
\end{itemize}

The rest of the paper is organized as follows. In \Cref{background}, we introduce the relevant theoretical background and present a brief literature survey. \Cref{threat} defines the threat model, describing the goals, knowledge, and capabilities of the attacker and the defender. \Cref{analysis} provides an analysis of the accuracy-privacy trade-off in the FL. Experimental results and discussions are provided in \Cref{results}. Finally, we conclude our work and discuss future research scope in \Cref{conclusion}.

\section{Background}\label{background}
In this section, we discuss the preliminary concepts and the background works on \textbf{deep ensembles and FL} along with various \textbf{federated aggregation techniques}. 

\subsection{Deep Ensembles and Federated Learning}

In the deep learning domain, deep ensembles are the predominant and heavily used method, where base models are initialized with random weights and trained on the same dataset \cite{rezaei2023accuracy}.
This approach consists of two steps: 1) training the base models on the same training dataset with random weights, and 2) fusing their prediction confidence by averaging to create the final output. This differs from classical ML ensembles, as diversity in deep ensembles primarily comes from the random initialization of base learners. Classical ensemble learning approaches, although adaptable, are seldom used in deep learning models due to their lower accuracy compared to deep ensembles \cite{lakshminarayanan2017simple}.

The consistent thread running through multiple studies establishes the prevalent assumption that FL inherently assumes the role of an implicit ensemble. Shi \textit{et al.}'s Fed-ensemble method \cite{shi2021fed}, capitalizing on model ensembling within the FL paradigm, stands out for its consistently superior performance compared to various FL algorithms. Similarly, Chen \textit{et al.}'s FedBE \cite{chen2020fedbe} introduces Bayesian model ensemble techniques to FL, illustrating the reliability of ensemble aggregation. Guha \textit{et al.}'s investigation into one-shot FL \cite{guha2019one} inadvertently reinforces the notion of implicit ensembling by demonstrating substantial improvements in Area Under the Curve (AUC) through ensemble learning and knowledge aggregation. Furthermore, Avdiukhin and  Kasiviswanathan's exploration of asynchronous FL \cite{avdiukhin2021federated}, while primarily focused on performance metrics, incidentally adds to the growing body of evidence suggesting that FL commonly embodies implicit ensemble characteristics. These recurrent findings collectively emphasize the widespread acknowledgment within the research community that FL naturally incorporates implicit ensemble mechanisms, contributing to improved generalization and overall performance.

\subsection{Aggregation Techniques in Federated Learning} 






In FL, the training process involves iterations between the central server and clients until a termination criterion is reached. This criterion may be a maximum number of iterations or a threshold for model accuracy. Once the process concludes, the central server converges the FL model ($\mathcal{M}$), which is then distributed to each of the clients in the system.

Aggregation techniques play a pivotal role in FL, a privacy-preserving ML paradigm. In FL, models are trained across decentralized devices, and aggregating their updates is essential to construct a global model. Common aggregation methods include Federated Averaging  \cite{mcmahan2017communication}, where local model updates are averaged, and Weighted Averaging, which considers the importance of each client's contribution based on factors such as data size or computation capability. These techniques enable the collaborative learning process while preserving the privacy of individual client data, making aggregation a critical aspect of the FL framework \cite{mothukuri2021survey}.

In 2017, McMahan \textit{et al.} introduced the Federated Averaging (FedAvg) algorithm using Stochastic Gradient Descent (SGD) \cite{mcmahan2017communication}. FedAvg involves the central server sharing global parameters and a model with a mini-batch of clients, which train the model on local data and share the updates. The global model is then created by averaging the weighted sum of the local updates at the centralized server, and training can be terminated based on a configurable round requirement. Followed by FedAvg, a set of improved averaging strategies were introduced in the literature, including Federated Proximal (FedProx \cite{li2020federated}), Federated Multi-task Adaptation (FedMA \cite{wang2020federated}), and Averaging with Hierarchical  Systems (HierFAVG \cite{liu2020client}).

Though each averaging strategy has its own merits, the classical FedAvg algorithm is relatively simple to implement and its reliability lies in its straightforward averaging of model updates, making it an ideal choice for fundamental investigations such as the one at hand.




\begin{figure}
    \centering
    \includegraphics[width=0.7\linewidth]{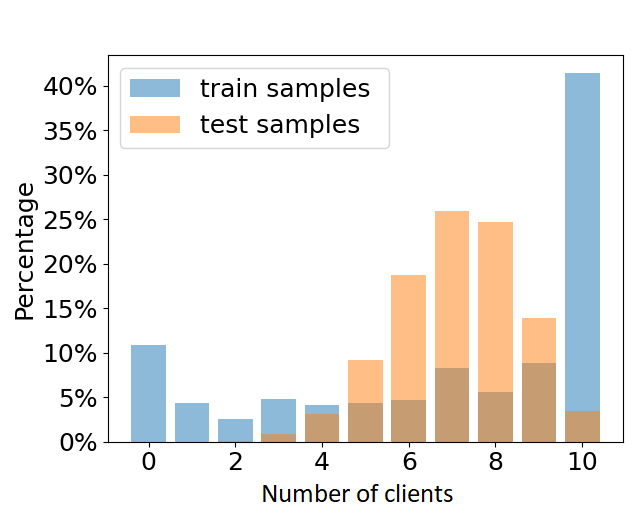}
    \caption{\textbf{Distinguishability between training and testing by measuring the agreement among FL clients on correct classifications}: The relative agreement among the $10$ FL clients in CIFAR10 using EfficientNet demonstrates a clearly apparent distributional shift between training and testing.}
    \label{fig:agrement}
\end{figure}


\begin{figure*}
    \centering
    \includegraphics[width=0.9\linewidth]{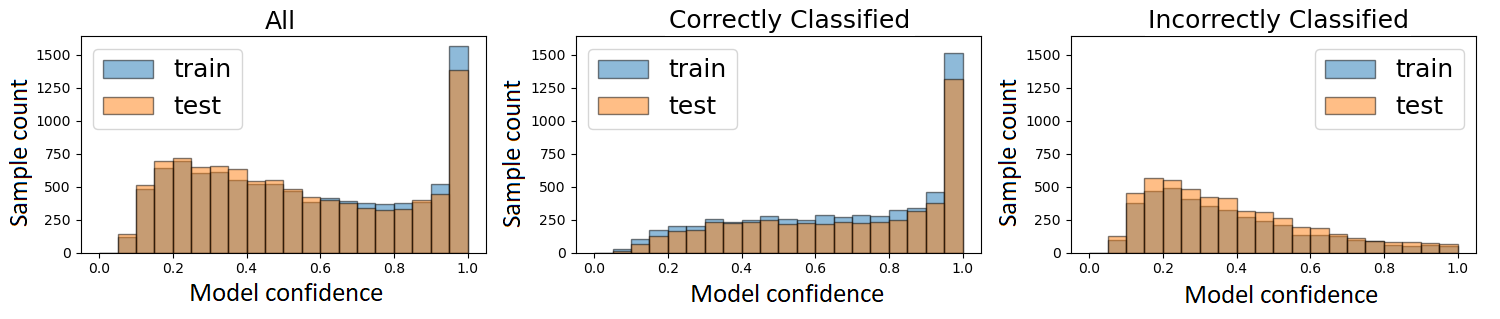}
    \caption{\textbf{Contrasting Confidence Between Correct and Incorrect Predictions}: The distribution of model predictions confidence on the CIFAR100 dataset highlights that models show high confidence when predictions are correct, whereas the confidence is significantly lower for incorrect predictions.}
    \label{fig:Confidence}
\end{figure*}

\subsection{Accuracy Issues of Membership Inference Attack}

The first introduced MIA on neural networks, known as Shokri's attack, involves training an attack classifier to predict membership status \cite{shokri2017membership}. Most MIA assumes that trained models exhibit higher confidence in member samples than non-member samples, using confidence levels as attack features \cite{mcmahan2017communication}, \cite{rezaei2021difficulty}. The attack classifier utilizes the victim model's prediction confidence as input, assuming greater confidence values for training samples compared to non-training samples. The attacker, with access to a dataset with a similar distribution, trains shadow models to replicate the victim model. Knowing the membership status of the data used to train shadow models, the attacker leverages this information to train the attack classifier.


With the exception of a few studies \cite{rezaei2021difficulty}, \cite{choquette2021label}, and \cite{rahimian2020sampling}, most prior research has focused on using prediction confidence to infer membership status. In \cite{rezaei2021difficulty}, the authors introduced MIA based on confidence levels, distance to the decision boundary, and gradients with respect to model weight and input, assuming white-box access. Interestingly, none of these methods outperformed confidence-based attacks. In \cite{choquette2021label}, two attacks based on input transformation and distance to the boundary were proposed in a limited information testing (Black-Box) setting. Similarly, \cite{rahimian2020sampling} introduced a sampling attack that randomly perturbs inputs to create a set of transformations, inferring membership status based on predicted labels. The rationale is that deep models are more robust on training samples, making perturbed samples less likely to be mislabeled. It demonstrated that Differentially Private Stochastic Gradient Descent (DP-SGD) effectively defends against the sampling attack, albeit at the cost of accuracy.

\textbf{Synthesis:} 
Existing research consistently highlights a trade-off between privacy and accuracy in ensemble learning, a challenge that resonates in the domain of FL. This study, aligning with this narrative, focuses on measuring the accuracy-privacy trade-off in FL. By employing MIA as a metric, the research aims to offer nuanced insights into the implications of privacy-preserving mechanisms in FL scenarios, addressing a critical gap in the current understanding.

\section{Threat Model} \label{threat}

In this section, we outline the fundamental threat model, detailing the defender and attacker perspectives \cite{rezaei2023accuracy}, \cite{shokri2017membership}, \cite{salem2018ml}.

\subsection{Defender's Perspective}

\textbf{Defender's  assumptions:}
In the context of FL, the defender utilizes the training dataset, using a disjoint subset of training samples for each client. A sample is classified as a member if it is used in at least one of the clients. 

The defender provides API access, returning prediction confidence values, especially in multi-class classification. Base models can be trained from scratch, allowing for the exploration of techniques that enhance privacy, modify the training process, and consider alternative fusing methods beyond the traditional confidence averaging associated with deep ensembles.

\textbf{Defender's  objectives:}
The defender's primary objective is to mitigate MIA within the FL framework while leveraging the accuracy benefits of ensemble learning. Mitigation of computational expenses is a primary concern throughout the training and inference processes. The investigation aims to achieve increased privacy protection with minimal loss to prediction accuracy within the FL paradigm.

\subsection{Adversary’s Perspective}

\textbf{Adversary’s knowledge:} We examine a challenging scenario where the adversary is limited to obtaining restricted information about the target model and knowledge concerning the distribution of the target dataset. The training goals and model architecture are universally shared among the FL participants, rendering this level of attacker knowledge realistic. However, the adversary is incapable of acquiring information regarding the global training process (whether centralized or federated) or the distribution of the training data among the clients.

\textbf{Adversary’s goals:} The adversary attempts to draw inferences or deductions of data from the initial training set. Subsequent to training, the attacker formulates an attack model that deduces private data through query-level access to the target model. The attacker refrains from modifying the parameters of the model and requires no additional information \cite{shokri2017membership}.

\textbf{Adversary’s  Capability:} The assumed attacker is an ostensibly honest yet inquisitive user, granted query access to the target model but without the ability to access its internal weights and gradients. Despite these limitations, the attacker leverages the provided query access to execute an MIA, attempting to discern if specific data points were part of the original training dataset.

\section{Accuracy-Privacy Trade-off in FL} \label{analysis}

In \cite{rezaei2023accuracy}, the authors present how in the context of ensemble learning there exists a similar tradeoff between accuracy and privacy. Moreover, they demonstrated that this is proportional to the number of ensembles. They demonstrated that ensemble learning increases accuracy at the cost of privacy. 

Membership inference attacks succeed in identifying the model reaction to the distributional difference between training and evaluation data. This is reflected in the difference between the evaluation confidence of the train and test samples of the model.  \Cref{fig:agrement} demonstrates how the different client's independent predictions agree when correctly classifying the training and test samples.  The figure highlights what percent of clients agree with the overall (FedAvg) correct classification.
From the figure, one can clearly distinguish the training and test samples from the agreement among the different clients. But thankfully for the defender, the adversary cannot directly exploit this information. However, MIA can leverage the effect of this agreement on the final model's prediction confidence. \Cref{fig:Confidence}(a) demonstrates the prediction confidence of EfficientNet trained on CIFAR100 dataset. The distributional shift between the training and test samples is clearly observed. \Cref{fig:Confidence}(b) and (c) show the same results, separated by correct and incorrect predictions. 

In the FL setting, despite forming an implicit ensemble, the number of clients is not correlated with the accuracy-privacy tradeoff. This is because, unlike in ensemble learning, the different clients in an FL setting do not have access to the entire dataset.  This phenomenon is investigated in the following section.

\begin{figure*}
    \centering
    \begin{subfigure}[b]{0.49\linewidth}
        \centering
        \includegraphics[width=\linewidth]{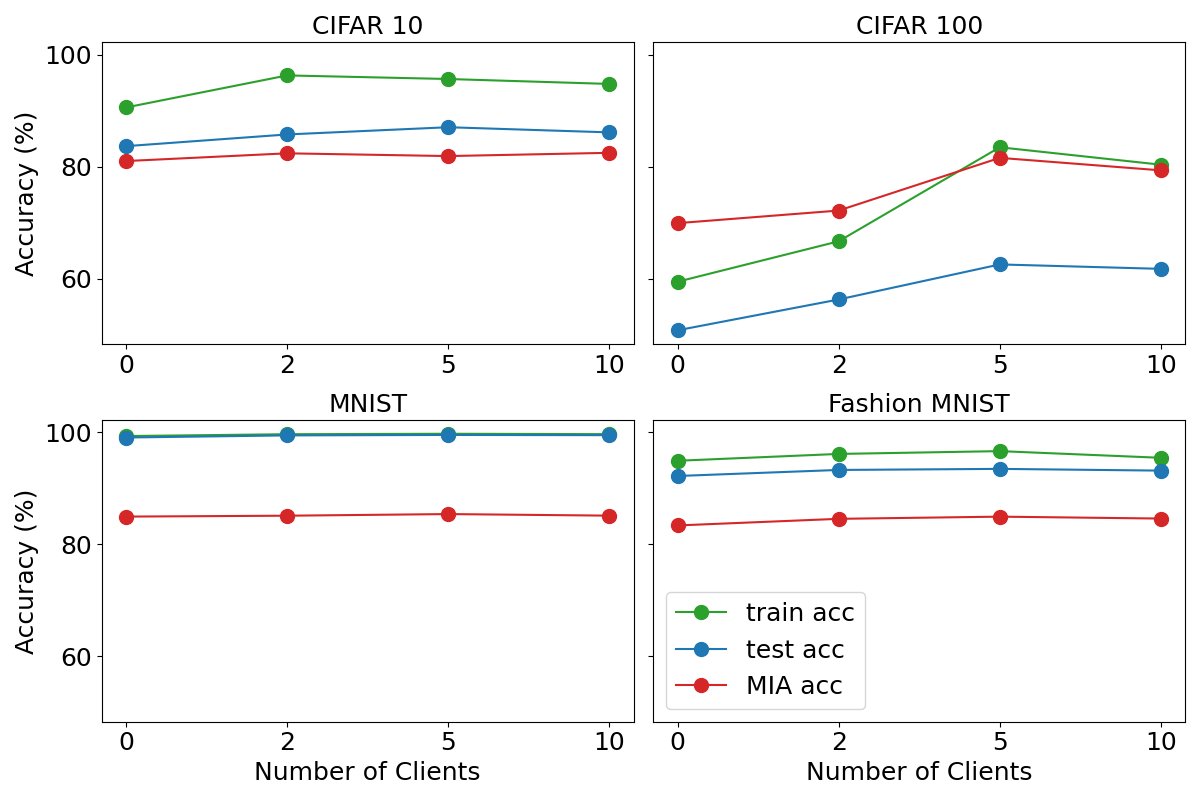}
        \caption{EfficientNet}
        \label{fig:overall_efficientnet}
    \end{subfigure}
    \hfill
    \begin{subfigure}[b]{0.49\linewidth}
        \centering
        \includegraphics[width=\linewidth]{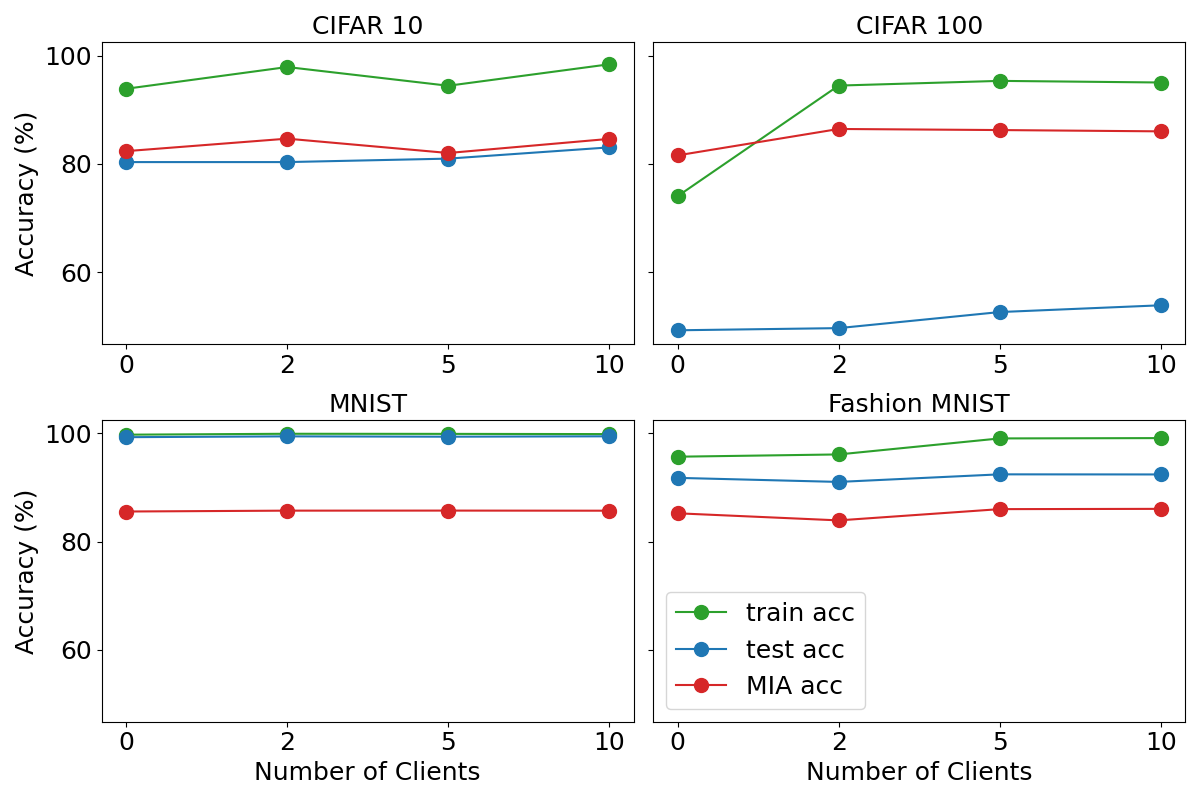}
        \caption{ResNet18}
        \label{fig:overall_resnet}
    \end{subfigure}
    \caption{ \textbf{Correlation of Accuracy and Privacy Across Datasets and Model Architecture}: Variation of accuracy and privacy with respect to the number of federated clients. While accuracy and privacy are strongly correlated across datasets and model architecture, the lack of correlation with the number of federated clients remains consistent.}
    \label{fig:overall_trends}
\end{figure*}

\section{Result Analysis and Discussion} \label{results}

In this section, we present our findings and discuss their implications. We provide various experimental results on the accuracy-privacy correlation and the effect of the trade-off for different datasets under various fusion strategies in FL.

\Cref{fig:overall_efficientnet,fig:overall_resnet} demonstrates that accuracy and privacy are strongly correlated in FL settings. We present how the training and latest accuracy (green and blue) are strongly correlated with the attacker's accuracy with the MI attack (red) for different datasets, such as CIFAR10 \cite{krizhevsky2009learning}, CIFAR100 \cite{krizhevsky2009learning}, 
from the Canadian Institute For Advanced Research, 
MNIST \cite{lecun1998mnist} from the Modified National Institute of Standards and Technology database, and Fashion-MNIST \cite{xiao2017fashion} from Zalando Research. The training data was evenly distributed among the clients in a disjoint way. We performed the experiment with EfficientNet and ResNet18\cite{he2016deep} models over the FL architecture with $2$, $5$, and $10$ clients.  In our experiment, the client $0$ is analogous to a centralized architecture.

\begin{figure}[!htb]
    \centering
    \fbox{
        \includegraphics[width=0.92\linewidth]{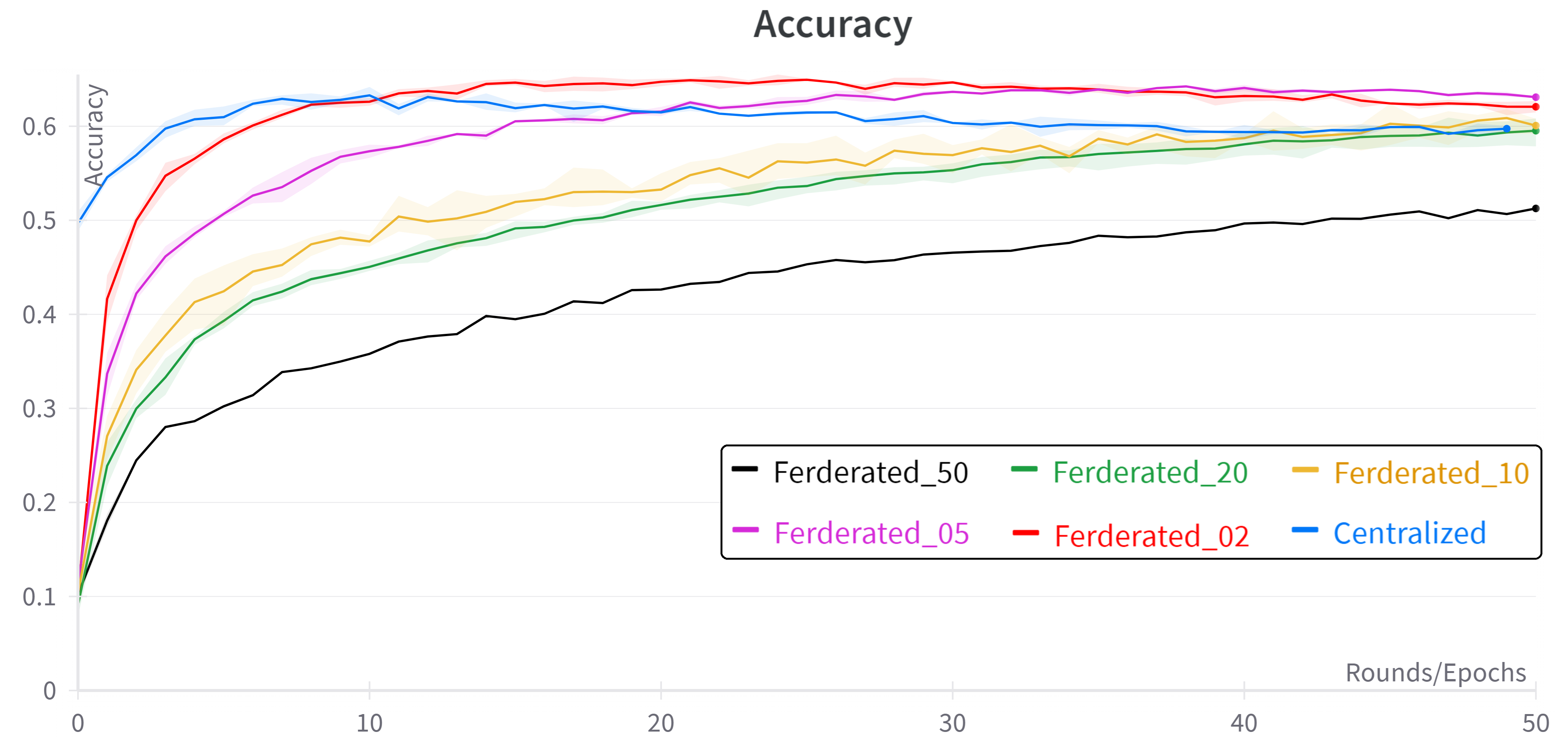}
    }
    \caption{ \textbf{Regularizing effect of the number of FL clients on accuracy}: with a larger number of clients, the models converge slower but tend to overfit less.}
    \label{fig:FL_acc}
\end{figure}

\Cref{fig:overall_efficientnet,fig:overall_resnet} further shows that there exists no strong correlation between accuracy (and privacy) and the number of clients. To investigate this phenomenon, we trained a small basic convolutional neural network (CNN) architecture over the CIFAR10 dataset, with up to 50 clients. We repeated each experiment $10$ times and presented the results in \Cref{fig:FL_acc}. We observed that the basic CNN model started overfitting due to its limited learning capacity. We did not implement early stopping for this experiment. This experiment demonstrated that FL provides a regularization effect on the model being trained. 
We see a distinct trend where the model convergence rate is slower when the number of federated clients is larger. However, for a larger number of clients, the model finally converges into a slightly higher accuracy value.
It's to be noted, that the improved accuracy over federated training was only observed for the basic CNN model, which did not have enough parameters to effectively learn the representation. For models such as EfficientNet, and ResNet18 such an increase is not observable.

In \cite{rezaei2023accuracy}, Rezaei \textit{et al.} proposed three methods to combine the ensemble model outputs that do not also make it easier for the MIA adversary to exploit. The three modes were to return a random base model, return the model that has the highest confidence, and return the model that has the highest agreed confidence. This approach works because the success of an MIA relies on a) confidence of the correct classification of the model, b) confidence of the incorrect classification of the model, and c) the level of agreement among the constituent models. Rezaei \textit{et al.} observed that in ensemble learning the third point, the level of agreement among the constituent models, becomes very significant. The three approaches proposed in \cite{rezaei2023accuracy} aim to disrupt the adversary's ability to exploit this information.

\begin{figure*}[htbp]
    \centering
    \begin{minipage}{0.495\textwidth}
        \begin{subfigure}[b]{\textwidth}
            \centering
            \includegraphics[width=\linewidth]{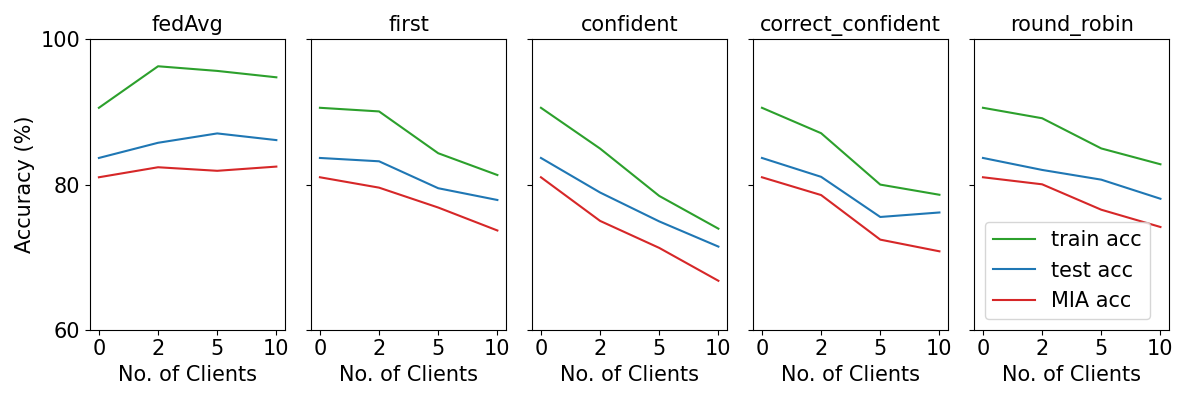}
            \caption{CIFAR10 with EfficientNet.}
            \label{fig:cifar10}
        \end{subfigure}
        
        \begin{subfigure}[b]{\textwidth}
            \centering
            \includegraphics[width=\linewidth]{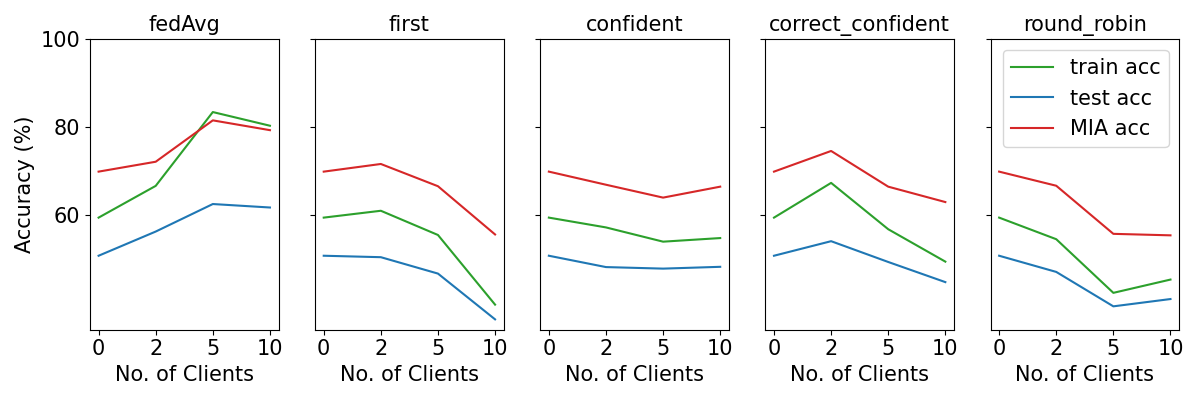}
            \caption{CIFAR100 with EfficientNet.}
            \label{fig:cifar100}
        \end{subfigure} 
        
        \begin{subfigure}[b]{\textwidth}
            \centering
            \includegraphics[width=\linewidth]{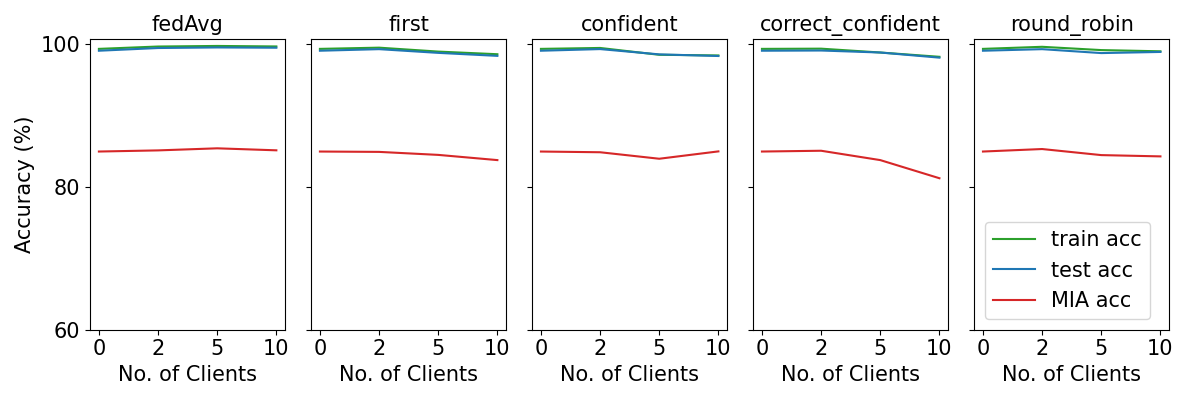}
            \caption{MNIST with EfficientNet.}
            \label{fig:mnist}
        \end{subfigure}
            
        \begin{subfigure}[b]{\textwidth}
            \centering
            \includegraphics[width=\linewidth]{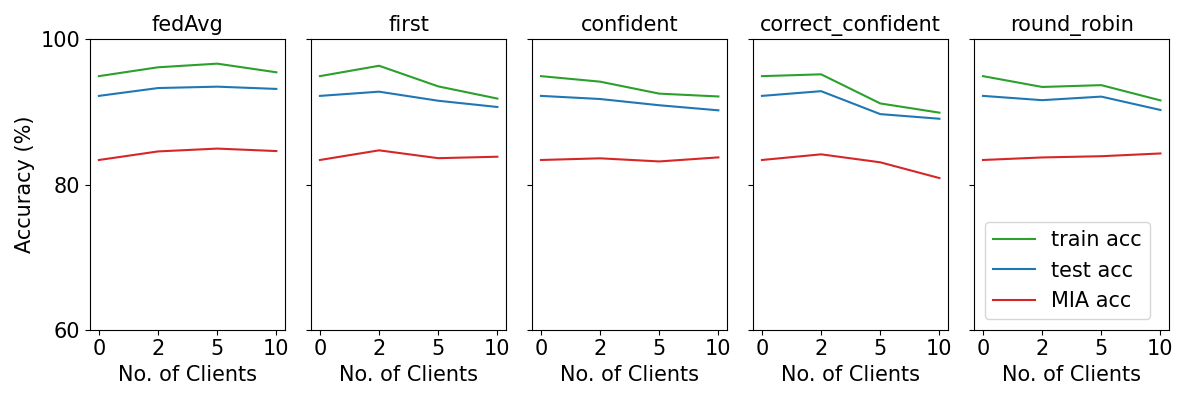}
            \caption{Fashion-MNIST with EfficientNet.}
            \label{fig:f_mnist}
        \end{subfigure}
    \end{minipage}
    \hfill
    \begin{minipage}{0.495\textwidth}
        \begin{subfigure}[b]{\textwidth}
            \centering
            \includegraphics[width=\linewidth]{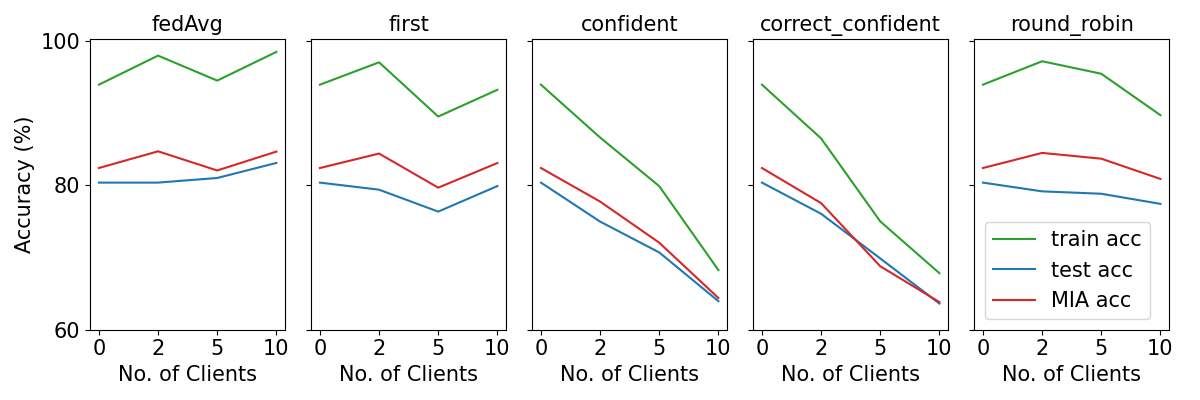}
            \caption{CIFAR10 with ResNet18.}
            \label{fig:cifar10_resnet}
        \end{subfigure}
        
        \begin{subfigure}[b]{\textwidth}
            \centering
            \includegraphics[width=\linewidth]{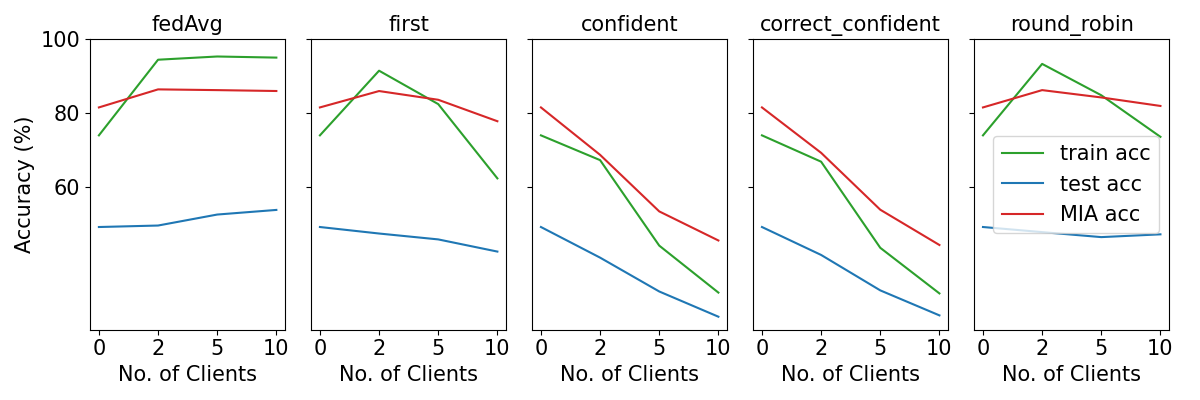}
            \caption{CIFAR100 with ResNet18.}
            \label{fig:cifar100_resnet}
        \end{subfigure}
    
        \begin{subfigure}[b]{\textwidth}
            \centering
            \includegraphics[width=\linewidth]{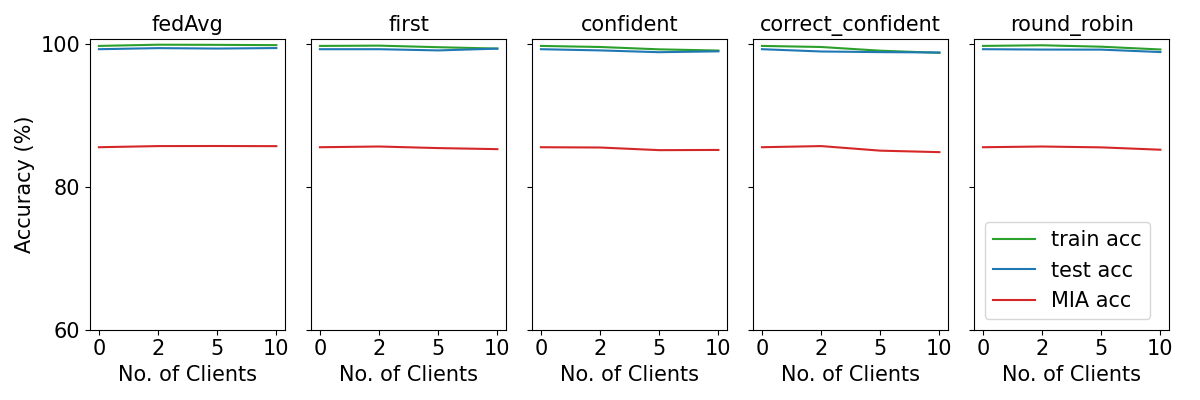}
            \caption{MNIST with ResNet18.}
            \label{fig:mnist_resnet}
        \end{subfigure}
        
        \begin{subfigure}[b]{\textwidth}
            \centering
            \includegraphics[width=\linewidth]{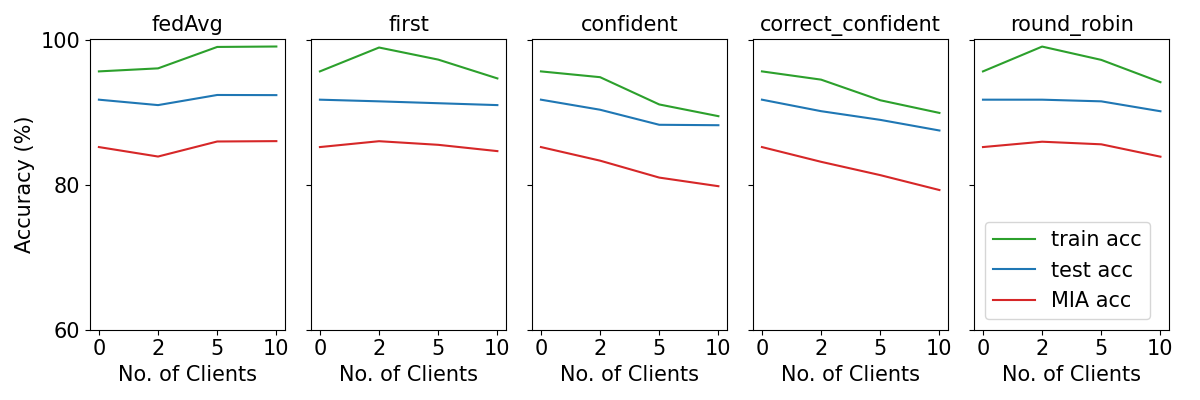}
            \caption{Fashion-MNIST with ResNet18.}
            \label{fig:f_mnist_resnet}
        \end{subfigure}
    \end{minipage}
    
    \caption{\textbf{Aggregation schemes for breaking the accuracy and privacy correlation proposed in \cite{rezaei2023accuracy} are ineffective in  FL}: across different datasets and model architectures the accuracy and privacy correlation persists regardless of the aggregation schemes.}
    \label{fig:averaging_schemes}
\end{figure*}

We investigated the effect of these recommendations in a FL setting. We re-defined the return of a random base model into two schemes, return the \textbf{first} model, and return the model identified by the round number $modulo$ number of clients, we call this the \textbf{round robin} scheme. We also tested the scheme where the most \textbf{confident} model and the \textbf{correct confident} model are returned.

 Considering EfficientNet architecture and CIFAR10, CIFAR100, MNIST, and Fashion-MNIST datasets, \Cref{fig:cifar10,fig:cifar100,fig:mnist,fig:f_mnist} demonstrate the effect of the privacy-preserving averaging schemes as compared to FedAvg.  
 Similarly, \Cref{fig:cifar10_resnet,fig:cifar100_resnet,fig:mnist_resnet,fig:f_mnist_resnet} demonstrate the effect of various privacy-preserving averaging schemes for ResNet18 architecture in FL environments. We can observe that, in terms of accuracy, each of the privacy-preserving averaging schemes performs worse than the FedAvg. This is a direct result of how these schemes are designed, as each of these schemes, at the end of the round, select the model from one single client to be retained. Thus the effective model is trained on approximately a dataset of $N/n$, where $N$ is the size of the original dataset and $n$ is the number of clients. This directly disadvantages settings with a large number of FL clients and accuracy gains through ensemble effects are canceled out. However, despite the drawbacks regarding the accuracy, when there is adequate data, and the effective model is trained to a high degree (e.g., MNIST), the proposed averaging schemes are remarkably effective in defending against MIA. 

\section{Conclusion and future Scope}\label{conclusion}

This study explores the accuracy-privacy trade-off in FL for MIA and investigates the applicability of confidence-based measures from deep ensembles in FL. Using different numbers of clients, datasets, and confidence-metric-based fusion techniques for various deep learning models, a clear accuracy-privacy trade-off was identified and analytically justified. However, empirical studies reveal a lack of a non-monotonic correlation between the number of clients and the accuracy-privacy trade-off.  The research identifies FL client data division as the primary contributor to accuracy loss. In the future, we aim to explore innovative approaches to enhance FL system privacy without compromising prediction accuracy.

\bibliographystyle{IEEEtran}
\bibliography{reference.bib}
\end{document}